\documentclass[10pt]{cai}

\begin{document}
\begin{center}


\title{Achieving Pareto Optimality using Efficient Parameter Reduction for DNNs in Resource-Constrained Edge Environment}
\maketitle

\thispagestyle{empty}

\begin{tabular}{cc}
Atah Nuh Mih\upstairs{\affilone,*}, Alireza Rahimi\upstairs{\affilone}, Asfia Kawnine\upstairs{\affilone}, Francis Palma\upstairs{\affilone,\affilfour}, \\Monica Wachowicz\upstairs{\affilone,\affiltwo}, Rickey Dubay\upstairs{\affilthree}, Hung Cao\upstairs{\affilone}
\\[0.25ex]
{\small \upstairs{\affilone} Analytics Everywhere Lab, University of New Brunswick, Canada} \\
{\small \upstairs{\affiltwo} RMIT University, Australia} \\
{\small \upstairs{\affilthree} Deparment of Mechanical Enigeerning, University of New Brunswick, Canada} \\
{\small \upstairs{\affilfour} SE+AI Lab, University of New Brunswick, Canada} \\
\end{tabular}
  
\emails{
  \upstairs{*}atah.mih@unb.ca
}
\vspace*{0.2in}
\end{center}

\begin{abstract}
This paper proposes an optimization of an existing Deep Neural Network (DNN) that improves its hardware utilization and facilitates on-device training for resource-constrained edge environments. We implement efficient parameter reduction strategies on Xception that shrink the model size without sacrificing accuracy, thus decreasing memory utilization during training. We evaluate our model in two experiments: Caltech-101 image classification and PCB defect detection and compare its performance against the original Xception and lightweight models, EfficientNetV2B1 and MobileNetV2. The results of the Caltech-101 image classification show that our model has a better test accuracy (76.21\%) than Xception (75.89\%), uses less memory on average (847.9MB) than Xception (874.6MB), and has faster training and inference times. The lightweight models overfit with EfficientNetV2B1 having a 30.52\% test accuracy and MobileNetV2 having a 58.11\% test accuracy. Both lightweight models have better memory usage than our model and Xception. On the PCB defect detection, our model has the best test accuracy (90.30\%), compared to Xception (88.10\%), EfficientNetV2B1 (55.25\%), and  MobileNetV2 (50.50\%). MobileNetV2 has the least average memory usage (849.4MB), followed by our model (865.8MB), then EfficientNetV2B1 (874.8MB), and Xception has the highest (893.6MB). We further experiment with pre-trained weights and observe that memory usage decreases thereby showing the benefits of transfer learning. A Pareto analysis of the models' performance shows that our optimized model architecture satisfies accuracy and low memory utilization objectives.   

\end{abstract}


\section{Introduction}




Recent advances in Artificial Intelligence (AI) and Machine Learning (ML) boost the proliferation of many AI-based applications and services. It is undeniable that new ML models such as large language models \cite{zhao2023survey} and diffusion models \cite{croitoru2023diffusion} are changing our lifestyles. However, these AI models require intensive computational resources such as CPU, GPU, memory, and network that only the cloud can offer. Cloud computing has been a key enabler of many new technologies \cite{cao2023fostering}, such as IoT, and AR/VR, by providing virtually unlimited resources, including on-demand storage and high computing power. By leveraging these advantages, many powerful AI models such as Segment Everything \cite{kirillov2023segment} are trained and deployed in the cloud. 

However, there are several drawbacks when implementing ML models that completely rely on the cloud. For example, the ML models running on the cloud will solely depend on the external infrastructure, leading to potential service interruptions and downtime if the cloud service provider experiences outages or technical issues. Moreover, cloud-based ML models may face latency and performance issues since their quality of services is closely tied to network quality. Unstable connectivity, bandwidth limitations, and network delays could cause many AI service interruptions.

These challenges motivate the need to rely on edge computing for training ML models. The rapid development of mobile chipsets and hardware accelerators has improved edge devices' computing power significantly \cite{shi2020communication}. This has led to a shift from deploying models in the cloud to the edge, where AI functionalities are diffused, converged, and embedded into resource-constrained devices in physical proximity to the users, such as micro data centers, cloudlets, edge nodes, routers, and smart gateways. However, this shift wave is only partially implemented and has not fully taken advantage of the power of edge computing. The literature highlights this since existing solutions only deploy the inference ML models at the edge \cite{sallang2021cnn, tsukada2020neural, cao2021edge}.

Training on the edge can prove beneficial in terms of variations between training and deployment environments and also address the viewpoint problem \cite{kukreja2019training}. However, the main challenge of training on the edge is the availability of computing resources, as modern deep learning architectures are designed to be computationally intensive. Although various lightweight deep learning models \cite{tan2021efficientnetv2,sandler2018mobilenetv2} have been proposed, they do not perform as well as their heavyweight counterparts. This leaves a research gap in developing deep learning architectures suitable for training on resource-constraint edge devices. We therefore aim to answer the following research question: 

\textit{``Can deep learning models be optimized to facilitate training at the edge with limited resources while maintaining high accuracy with less resource consumption?"}


In this paper, we optimize an existing deep neural network architecture with state-of-the-art performance to improve its on-device training in an edge environment. We adopt strategies described by Iandola et al. \cite{iandola2016squeezenet} that enable low model size while preserving high accuracy. For the existing deep learning model, we choose Xception \cite{chollet2017xception} as a backbone to integrate the strategies and implement two experiments to evaluate its training performance. We compare the results against the original Xception as baseline and also against lightweight models, EfficientNetV2B1 and MobileNetV2.




The main contributions of this work are as follows:
\begin{enumerate}
    \item We present an optimization of existing deep neural networks, which facilitates efficient hardware utilization for training in resource-constrained edge environments. 
    \item We implement this optimization on the Xception architecture and evaluate its performance in terms of accuracy, memory usage, and inference latency on Caltech-101 and a PCB defect detection task.
    \item We explore the benefits of transfer learning on the resource utilization of models by comparing the performance of pretrained models vs non-pretrained models.  
\end{enumerate}


Our paper is structured in the following order:
Section 1 introduces the background of Edge AI, its challenges, and the problem to be solved. 
Section 2 discusses relevant related work involving model optimization and machine learning with edge devices. 
Section 3 describes our implementation of a memory-efficient optimization using efficient parameter reduction. 
In Section 4, we implement our proposed architecture on an edge device and experiment on Caltech-101 image classification and a PCB defect detection task, and present the results of our finding. 
We analyze the results of our experiment and conclude our paper in Section 5. 
\section{Related Work}
This section presents a summary of the related works in this research area. We first discuss relevant literature on model optimization techniques and then proceed to explore literature on machine learning with edge devices.

\subsection{Model Optimization using Post-Training Quantization}
The most common model optimization methods involve the use of compression techniques for improved hardware performance. One such method is post-training quantization (PTQ), which includes techniques to reduce hardware utilization and model size. Post-training quantization converts a pre-trained FP32 network into a fixed-point network through various quantization methods while omitting the original training pipeline \cite{nagel2021white}.    

Post-training quantization has been widely used as a model compression technique. Victor Habi et al. \cite{victorhabi2021hptq} proposed a hardware-friendly post-training quantization (HPTQ) framework that achieves hardware efficiency by combining several quantization techniques such as channel equalization, threshold selection, per channel quantization, shift negative correction, and bias correction. They achieve a peak quantization accuracy of 75.018\% on ImageNet with ResNet50. Banner et al. \cite{banner2019post} proposed 4-bit PTQ that targets both weight and activation quantization, and they proposed methods for minimizing quantization error. Wu et al. \cite{wuhao2020integer} proposed an 8-bit quantization approach that maintain comparable accuracy as the FP32 baseline on hard-to-quantize networks such as MobileNets and BERT. 

Several other PTQ have been proposed such as loss-aware post-training quantization \cite{nahshan2021loss}, post-training piecewise linear quantization \cite{fang2020post}, and adaptive rounding for post-training quantization \cite{pmlr-v119-nagel20a}. The challenge of these compression techniques is attributed to the iterative training process that makes it difficult to use complex optimization algorithms. As such model compression techniques are often used for inference as they make training difficult to speed up \cite{dengb2020model}, and we therefore exclude PTQ for our approach.

\subsection{Model Optimization with Neural Architecture Search}
Model optimization can be equally conducted as a neural architecture search (NAS) process. With neural architecture search, a controller decides the best architecture for a given task by using search objectives such as accuracy, latency, and resource utilization. NAS as an optimization technique determines the best model given specific objectives to attain. Various NAS approaches exist such as NAS-RL \cite{nasRL}, ENAS \cite{enas}, DARTS \cite{darts}, efficient architecture search \cite{cai2018efficient}, and PNAS \cite{pnas}. 

Although NAS often results are often successful, the search process is usually long and resource intensive. Searching from scratch fails to take advantage of the existing neural architectures and overlook the neural architecture design expertise that already exists \cite{renp2021comprehensive}. As such, NAS has been explored by using existing model architectures as baselines that define the search space. Li et al. \cite{liy2023pareto} used a ResNet-18 backbone for their search process, while Lyu et al. \cite{lyub2021resource} used MobileNetV2. The limitation of these works is their choice of lightweight models, which already have a significantly lower accuracy than heavyweight models. By re-designing a heavyweight architecture, we explore new ways of improving resource efficiency of the architecture without losing significant accuracy. 

Pre-defining search patterns can help guide the process towards better architectural decisions and therefore shorten the process by constraining the search space. Shortening the search process is a suitable trade-off to consider for resource-constraint devices. Establishing these patterns, however, require prior knowledge of architectural design to guarantee the success of the pattern. In this paper, we focus on establishing the success of an optimization method (which could become a search pattern) and therefore deviate from using NAS headfirst. 

\subsection{AI Models at the Edge}
Many studies have implemented artificial intelligence on edge. Nikouei et al. \cite{nikouei2018real} developed a lightweight CNN (L-CNN) using depthwise separable convolution and a Single Shot Multi-Box Detector (SSD) for human object detection and deployed the model on an edge device, Sallang et al. \cite{sallang2021cnn} deployed a MobileNetV2-based SSD on a Raspberry Pi 4 for smart waste management; and Sreekumar et al. \cite{sreekumar2018real} designed a real-time traffic pattern collection method using YOLOv2 deployed on an edge device.

Beyond the use of edge devices for deploying machine learning models, other authors explored performing on-device training. Kukreja et al. \cite{kukreja2019training} proposed using a student-teacher model for training, where a teacher model is trained on an object and used to update the dataset with different viewpoints on which student models are trained. They also discuss the use of checkpointing to reduce the memory consumption of the training process. Tsukada et al. \cite{tsukada2020neural} proposed an On-device Learning Anomaly Detector (ONLAD), which combines sequential learning with semi-supervision and an autoencoder to reduce computational cost. They developed a hardware implementation of their method called ONLAD Core, on which they performed on-device training. 

Similar to these works, we emphasize on-device training by implementing and training our models on the edge. However, we differ from these approaches by optimizing a deep neural network to make it lightweight and computationally efficient enough to train on the edge. This enables us to leverage the well-proven architecture to obtain higher accuracy than other lightweight models.




\section{Proposed Approach and Implementation}
Deep learning models can be optimized through any of the following techniques: quantization, neural architecture search, pruning, knowledge distillation, and compact network design. Compact network design (CND) use of hardware-efficient techniques for designing neural network architectures that enable improved training efficiency for models. These techniques include depthwise separable convolutions, residual connections, and intermediate data encoding \cite{Shuvo2023}.

\begin{figure*}[h!]
    \centering
    \includegraphics[width=1\linewidth, height=5.5cm, keepaspectratio]{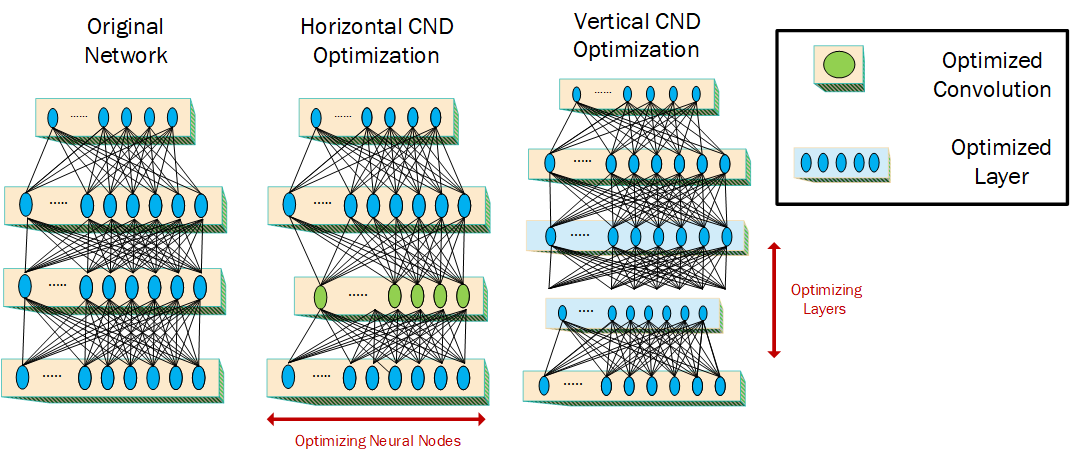}   
  \caption{Optimizing a model using Compact Network Design.}
        \label{fig:methodology}
\end{figure*}

The use of compact network design can be extended to existing DL models for optimized hardware performance. As such, we propose two paradigms for compact network design in the optimization of existing DNNs as illustrated in Figure \ref{fig:methodology}: (i) horizontal compact network design optimization; and (ii) vertical compact network design optimization. Horizontal CND optimization preserves the network architecture to leverage its proven success, but integrates optimized convolutions. It aims to find the optimal number of channels, types of filters, and other in-layer factors that achieve both objectives of high performance and better resource utilization. This enables the optimized model achieve similar performance to the original architecture, but a better hardware utilization. Vertical CND optimization, on the other hand, relies on the overall network architecture as a template upon which an optimized network can be implemented. An example of such an optimization is the design of the GRU as an efficient simplification of the LSTM \cite{Shuvo2023}.  

In this paper, we optimize the DNN model using the horizontal reduction approach (as depicted in Figure \ref{fig:methodology}). Particularly, we focus on optimizing Xception \cite{chollet2017xception}. We use an efficient parameter reduction approach described by \cite{iandola2016squeezenet} which we will discuss in the next section. For each setting, we observe the training performance and the efficiency metrics to ensure that each objective is achieved without a substantial cost to the other objective.



\begin{figure*} [h!]
    \centering
    \includegraphics[width=1\linewidth]{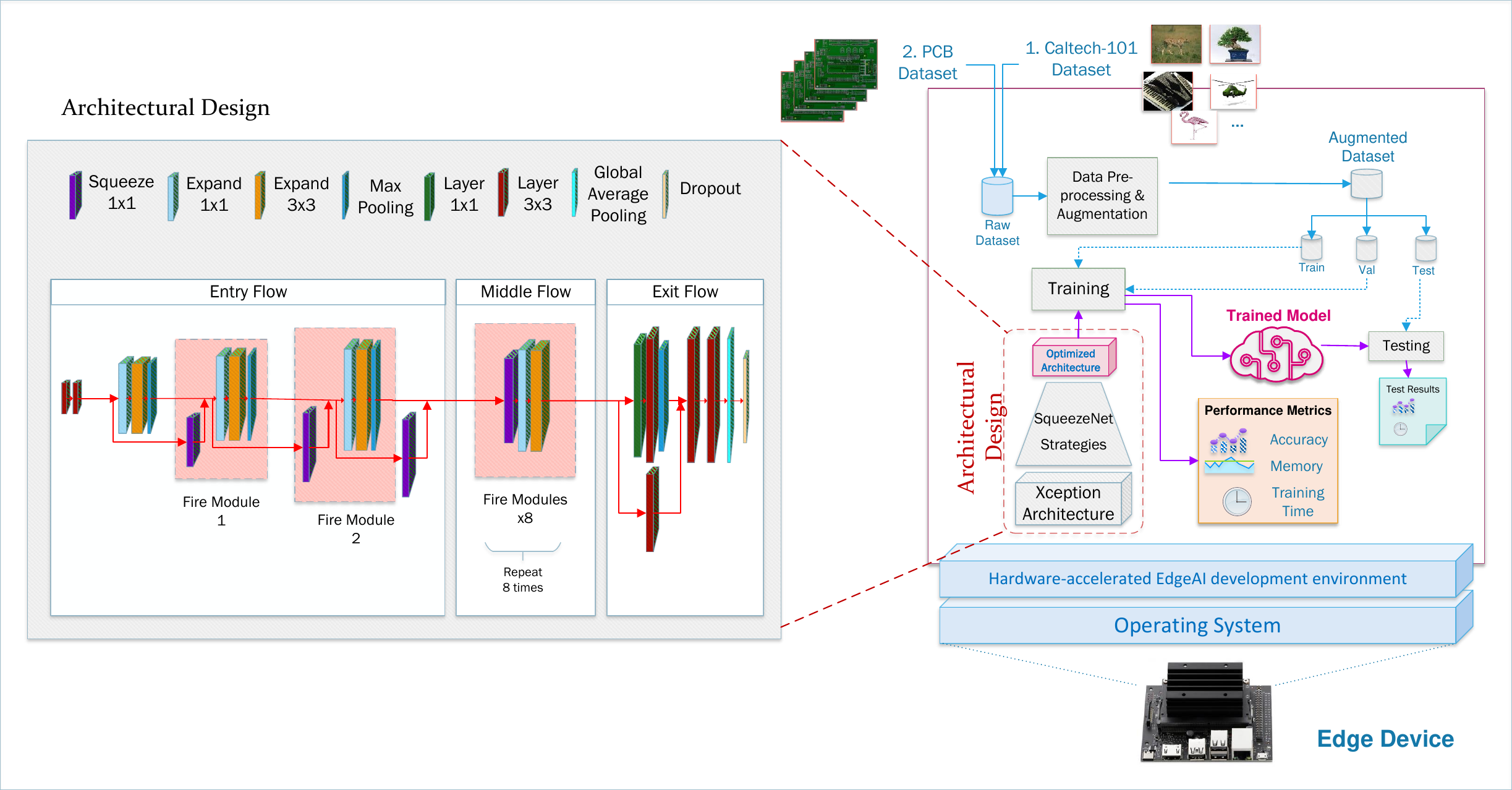}   
  \caption{Implementation of an optimized DNN.}
        \label{fig:implementation}
\end{figure*}

An overview of our implementation is shown in Figure \ref{fig:implementation}.

\subsection{Xception}
The Xception architecture \cite{chollet2017xception} consists of a linear stack of depthwise separable convolutions. In this architecture, 36 convolutional layers are assembled into 14 modules with residual connections around them, except for the first and last modules. These modules are divided into an Entry flow, a Middle flow, and an Exit flow. Depthwise separable convolution and residual connections are two factors that have been reported to be essential in the design of lighter and more efficient network architectures \cite{nikouei2018real, wangX2020convergence, xuR2022efficient}. 

Xception's 79.0\% Top-1 Accuracy on ImageNet indicates the success of the model's macro-architectural design. Our goal is to optimize this macro-architecture by using horizontal compact network design methods to improve its resource efficiency. We employ design methods explored in SqueezeNet to achieve this objective.


\subsection{SqueezeNet}\label{meth:squeezenet}
Iandola et al \cite{iandola2016squeezenet}, proposed SqueezeNet as a lightweight CNN model with 50 times fewer parameters than AlexNet but with a comparable performance. They describe three strategies to reduce the number of parameters of a CNN architecture: Strategy (1) Replacing 3x3 filters with 1x1 filters; Strategy (2) Decreasing the number of input channels to 3x3 filters; and Strategy (3) Downsampling late in the network so that convolutional layers have large activation maps.

The authors also present a \textit{fire module} consisting of a \textit{squeeze} layer (containing only 1x1 convolutional filters), followed by two \textit{expand} layers (a mix of 1x1 and 3x3 convolutional filters). 


Using these design considerations, they propose a SqueezeNet architecture containing a single convolutional layer, followed by 8 \textit{fire modules}, and a final convolutional layer.   

\subsection{Optimizing Xception}
Xception's modular architecture makes it flexible for optimization and thus motivates our choice to optimize the architecture. Our approach includes modifying the Xception network to include Strategies (1) and (2) of SqueezeNet which aim to reduce the number of parameters while preserving accuracy. We maintain the original macro-architectural design of Xception (i.e. Entry flow, Middle flow, and Exit flow), but alter the micro-architecture (i.e. varying the number of filters and channels). 

All first 3x3 filters in Separable Convolution layers are replaced with 1x1 filters, thereby satisfying Strategy (1). This technique reduces the number of filters nine-fold and subsequently reduces the number of parameters.    

The number of parameters of a layer is defined by: 
\begin{equation}
    \omega = N_{channels} * M_{filters} * \Psi_{filter} 
\end{equation}

where \\ 
    $\omega$ is the number of parameters, \\
    $N_{channels}$ is the number of channels \\
    $M_{filters}$ is the number of filters \\
    $\Psi_{filter}$ is the dimensions of the filter, e.g. 1x1 or 3x3

\vspace{2mm}
Strategy (1) already explores reducing the number of filters by replacing 3x3 filters with 1x1 filters. By implementing Strategy (1), we have a fixed number of filters with fixed filter sizes. Equation 1 becomes a linear relationship between the number of parameters and the number of channels. This relationship is exploited by Strategy (2) which proposes decreasing the number of input channels into a layer. 

We reduce the number of channels in the Entry Flow and Middle Flow to ensure that fewer channels are fed into subsequent layers of the network. The \textit{fire module} described in Section \ref{meth:squeezenet} also provides an implementation of Strategy (2). The \textit{squeeze} layer of the \textit{fire module} reduces the number of channels that are fed into the final \textit{expand} layer (3x3 filters).  

We follow the \textit{fire module} pattern in our implementation by modifying the layers in the Entry and Middle flows, such that there are \textit{fire modules} with a \textit{squeeze} and two \textit{expand} layers. A \textit{squeeze} layer has 1x1 filters, the first \textit{expand} layer has 1x1 filters, and the second \textit{expand} layer has 3x3 filters. The following relationship is preserved in the architecture: 

\begin{equation} \label{eqn:fire}
    s_{1x1} < e_{1x1} + e_{3x3}
\end{equation}
where $s_{1x1}$ is the number of filters in the \textit{squeeze} layer, $e_{1x1}$ is the number of filters in the first \textit{expand} layer, and $e_{3x3}$ is the number of filters in the last \textit{expand} layer.    

The implementation of a few pooling layers early in the network maintains large activation maps as the layers progress. These are later downsampled late in the network in accordance with Strategy (3).

\section{Experiments and Results}
Our edge environment consists of an A203 Mini PC built with Nvidia's Jetson Xavier NX 8GB module, 128GB SSD, and a pre-installed JetPack 5.0.2 on Ubuntu 20.04. The A203 Mini PC is a powerful edge computer that brings AI to the edge. With up to 21 TOPS and an integrated GPU, it provides AI computational capabilities for smart cities, industrial automation, and smart manufacturing. We implement two experiments to validate our results: Caltech-101 image classification and a PCB defect detection.

\subsection{Experiment 1: Image Classification with Caltech-101} \label{4:caltech}
The first experiment evaluates our model performance on image classification with Caltech-101 \cite{caltech}. The dataset consists of 101 object classes, including "airplanes", "elephant", "saxophone", and several others. Each class of objects contains between 40 to 800 images and a total of 9,146 images.

\begin{figure}[t!] 
    \centering
    \subfloat[Training Accuracy\label{fig_caltech_acc}]{%
    \includegraphics[width=0.5\linewidth, height=6cm, keepaspectratio]{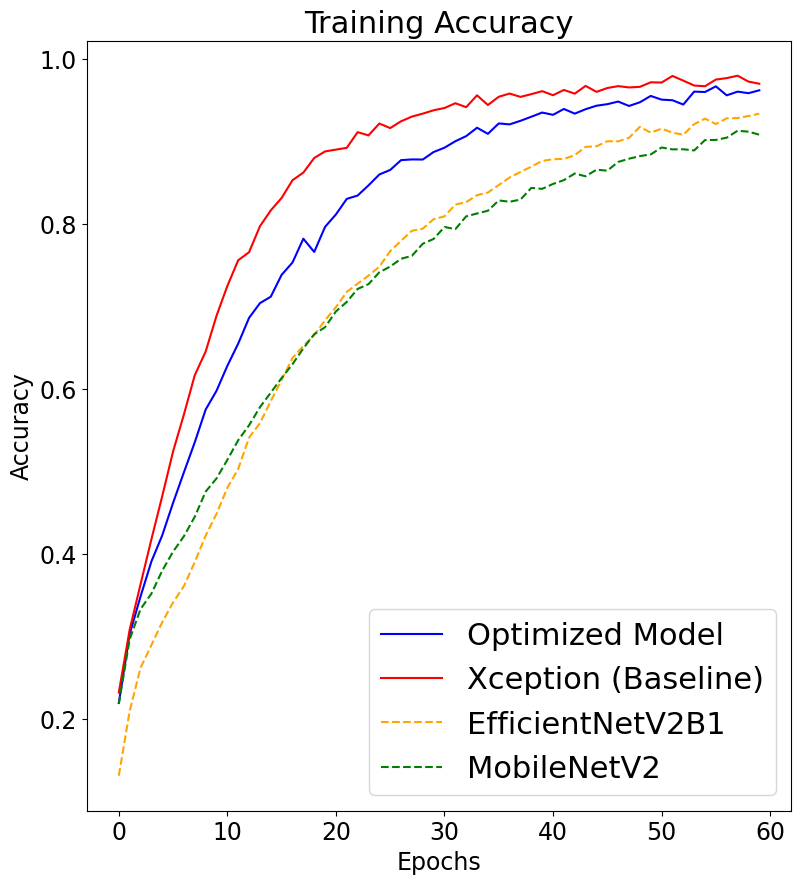}}
  \subfloat[Memory Usage\label{fig_caltech_mem}]{%
        \includegraphics[width=0.5\linewidth, height=6cm, keepaspectratio]{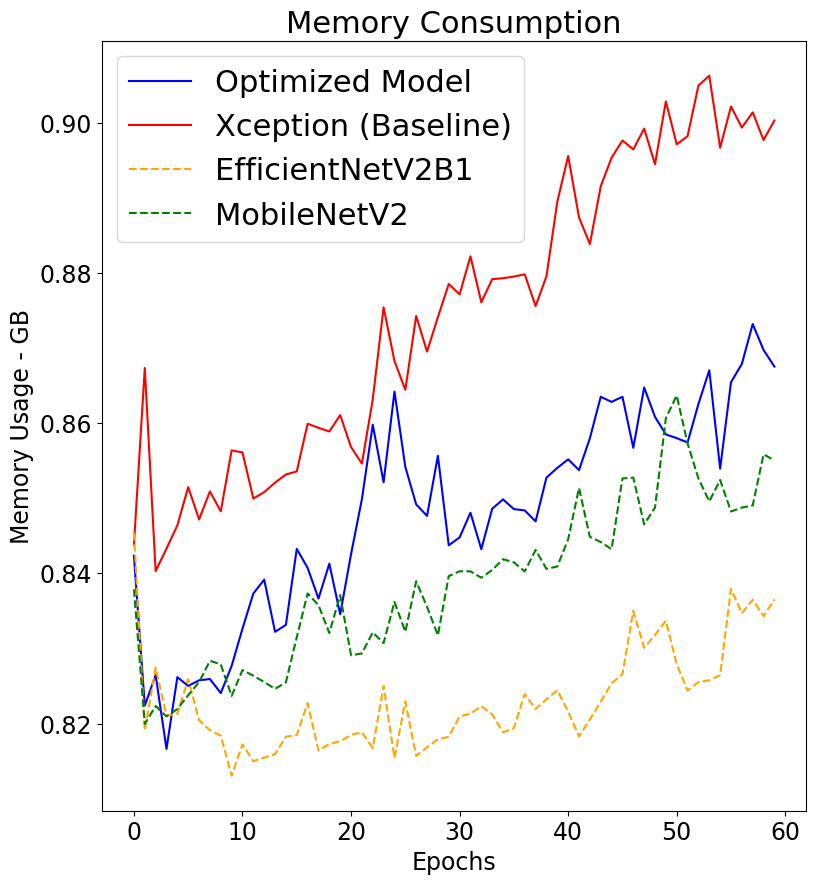}}    
  \caption{Accuracy and Memory Usage Patterns During Training (compared with original Xception architecture as baseline and other lightweight models, EfficientNetV2B1 and MobileNetV2)}
        \label{fig:caltech_train}
\end{figure}

Our results are presented in Figure \ref{fig:caltech_train} and Table \ref{tab:caltech_compare}. We make a primary comparison of our proposed model against the original Xception architecture, but also include the results for the lightweight models MobileNetV2 and EfficientNetV2B1. 

\begin{table*}[t!]
    \centering
    \caption{Comparison of model performance on Caltech-101. The base comparison is made with the original Xception model.  $\uparrow$ represents an increase in a metric compared to  Xception while $\downarrow$ represents a decrease in the metric.}
    \resizebox{14cm}{!}{%
\begin{tabular}{l|l|llllll}
\hline \hline
\cline{2-8}
 & Model & \#Params & Train Acc & Test Acc & \begin{tabular}[c]{@{}l@{}}Avg Mem.\\ per Epoch\end{tabular} & \begin{tabular}[c]{@{}l@{}}Avg Time \\ per Epoch\end{tabular} & \begin{tabular}[c]{@{}l@{}}Avg Inf.\\ Time\end{tabular} \\ \hline
\multirow{2}{*}{\begin{tabular}[c]{@{}l@{}}\textcolor{magenta}{Baseline}\\ \textcolor{magenta}{Comparison}\end{tabular}} & \textcolor{blue}{Optimized Model} & \textcolor{blue}{15.8M $\downarrow$} & \textcolor{blue}{96.16\%$\downarrow$} & \textcolor{blue}{76.21\%$\uparrow$} & \textcolor{blue}{847.9MB$\downarrow$} & \textcolor{blue}{523.88s$\downarrow$} & \textcolor{blue}{465ms$\downarrow$} \\
 & Xception (Original) & 21.1M & 96.95\% & 75.89\% & 874.6MB & 702.19s & 520ms \\ \hline
\multirow{2}{*}{\begin{tabular}[c]{@{}l@{}}\textcolor{cyan}{Other Lightweight} \\\textcolor{cyan}{Models}\end{tabular}} & EfficientNetV2B1 & 7.1M & 93.32\% & 30.53\% & 823.0MB & 381.44s & 383ms \\
 & MobileNetV2 & 2.4M & 90.79\% & 58.11\% & 838.6MB & 301.12s & 306ms \\ \hline
\end{tabular}
}   
    \label{tab:caltech_compare}
\end{table*}

From the results, it can be observed that EfficientNetV2B1 and MobileNetV2 had high training accuracies but very low test accuracies indicating that they overfit the training data. This was not observed in both Xception and in our proposed architecture. Due to this overfitting, we focus our comparison on the test accuracy as it is more indicative of the model's generalizability. EfficientNetV2B1 had the lowest test accuracy (30.53\%), MobileNetV2 had a test accuracy of 58.11\%, Xception had a test accuracy of 75.89\% and our proposed model had the highest test accuracy 76.21\%. This comparison shows the shortcomings of lightweight models in terms of accuracy as both EfficientNetV2B1 and MobileNetV2 had very low test accuracies. 

For memory consumption, Xception stood out from the other models, with the highest average memory consumption of 874.6MB. The consumption pattern was also observed in Figure \ref{fig_caltech_mem}. Meanwhile, EfficientNetV2B1 had the lowest average memory usage of 823.0MB and MobileNetV2 had 838.6MB. Our proposed model had an average memory consumption of 847.9MB, which was close to that of MobileNetV2 and significantly lower than that of Xception. Xception's high memory usage shows the limitation of heavyweight models in this aspect.

\subsection{Experiment 2: PCB Defect Detection} \label{4:transferd2}
This experiment evaluates our model performance on a PCB defect detection task as described by Nuh et al. \cite{nuh2023transferd2}. The PCB Defect Dataset \cite{huang2019pcb} contains 1386 images divided into 6 types of defects (Missing Hole, Open Circuit, Mouse Bite, Spurious Copper, Short, and Spur) in this dataset. The technique in \cite{nuh2023transferd2} includes an augmentation technique that provides an augmented dataset of 22,000 images. It also explores defect detection as a binary classification problem, which reduces the complexity of implementation and allows a more straightforward analysis of resource consumption. As the technique leverages transfer learning, we make two comparisons of model performance in this experiment. In the first setting, we implement the models without any pre-trained weights, and  implement the pre-trained models from Section \ref{4:caltech} in the second setting. The results of both settings are presented in Table \ref{tab:pcb_results}.

\begin{table*}[t!]
    \centering
    \caption{Comparing PCB Defect Detection task with base models vs using models pre-trained on Caltech-101. $\uparrow$ represents an increase in the pre-trained model's metric against its base model, while $\downarrow$ represents a decrease.}
    \resizebox{14cm}{!}{%
\begin{tabular}{l|llll|llll}
\hline
& \multicolumn{4}{c|}{\textcolor{orange}{\textbf{Not Pre-trained}}}                                   & \multicolumn{4}{c}{\textcolor{orange}{\textbf{Pre-trained on Caltech-101}}} \\ \hline
\hline
Model &
  \begin{tabular}[c]{@{}l@{}}Train\\ Acc.\end{tabular} &
  \begin{tabular}[c]{@{}l@{}}Test\\ Acc.\end{tabular} &
  \begin{tabular}[c]{@{}l@{}}Avg\\ Mem.\\ (MB)\end{tabular} &
  \begin{tabular}[c]{@{}l@{}}Avg\\ Inf.\\ Time\end{tabular} &
  \begin{tabular}[c]{@{}l@{}}Train \\ Acc.\end{tabular} &
  \begin{tabular}[c]{@{}l@{}}Test\\ Acc.\end{tabular} &
  \begin{tabular}[c]{@{}l@{}}Avg\\ Mem.\\ (MB)\end{tabular} &
  \begin{tabular}[c]{@{}l@{}}Avg\\ Inf.\\ Time\end{tabular} \\ \hline
\textcolor{blue}{Optimized Model}             & \textcolor{blue}{\textbf{97.81\%}} & \textcolor{blue}{\textbf{90.30\%}} & \textcolor{blue}{865.8} & \textcolor{blue}{465ms} & \textcolor{blue}{70.71\% $\downarrow$}    & \textcolor{blue}{69.80\% $\downarrow$}   & \textcolor{blue}{833.7 $\downarrow$}   & \textcolor{blue}{446ms $\downarrow$ }   \\
Xception (Original)         & 97.66\%          & 88.10\%          & 893.6          & 519ms          & \textbf{70.84}\% $\downarrow$      & \textbf{71.00}\% $\downarrow$      & 831.5 $\downarrow$     & 509ms $\downarrow$    \\
EfficientNetV2B1 & 88.87\%          & 55.25\%          & 874.8          & 346ms          & 65.96\% $\downarrow$   & 67.40\% $\uparrow$   & 828.7 $\downarrow$   & 335ms $\downarrow$    \\
MobileNetV2      & 96.13\%          & 50.50\%          & \textbf{849.4 }         & \textbf{295ms}          & 67.89\% $\downarrow$   & 69.05\% $\uparrow$    & \textbf{818.2} $\downarrow$   & \textbf{295ms} $\downarrow$    \\ \hline
\end{tabular}
}   
    \label{tab:pcb_results}
\end{table*}

\subsubsection{Non Pre-trained Models}
From Table \ref{tab:pcb_results}, we observed that our model has the highest training accuracy (97.81\%) and test accuracy (90.30\%). Similarly, Xception had a high training accuracy (97.20\%) and the second best test accuracy (88.10\%). EfficientNetV2B1 and MobileNetV2 overfit with high training accuracies but low test accuracies. MobileNetV2 had the least average memory usage of 849.4MB followed by our proposed model with 865.8MB. EfficientNetV2B1 had an average memory utilization of 874.8MB and Xception had the highest usage (893.6MB). MobileNetV2 also had the least inference time (295ms) followed by EfficientNetV2B1 (346ms), then our proposed model (465ms), and finally Xception (519ms).  

These results prove that our model retains its high accuracy on both training and testing, reflecting its generalizability. It maintains a comparable accuracy with Xception but has a better memory usage than the latter.

\subsection{Pre-trained Models}
A different picture was observed when using pre-trained models. The lightweight models (EfficientNetV2B1 and MobileNetV2) that previously overfit, had more stable training and test performance. EfficientNetV2B1 had a training accuracy of 65.96\% and a test accuracy of 67.40\%. MobileNetV2 had a training accuracy of 67.89\% and a test accuracy of 69.05\%. Our model had a training accuracy of 70.71\% and a test accuracy of 69.80\%. Xception had both the best training (70.84\%) and test accuracy (71.00\%). 

In this setting, the lightweight models had a more stable training performance and did not overfit the training data as compared to their non pre-trained settings. This performance improvement was contrary to that of our model which had a lower training accuracy of 70.71\% and a lower test accuracy of 69.80\%. Xception's training accuracy also decreased from 97.20\% to 70.84\% and its test accuracy decreased from 88.10\% to 71.00\%. 

MobileNetV2 maintained the least average memory utilization (818.2MB). EfficientNetV2B1 used less memory (828.7MB) than our model (833.7MB). Xception also had improved memory usage (831.5MB) which was lower than ours. We can also observed that the accuracies of all the models were very comparable when using pre-trained weights and they have a better memory usage. 

\section{Discussion and Conclusion}
In this paper, we explored a neural network optimization to improve hardware efficiency while maintaining high accuracy. It may seem trivial that reducing the number of parameters would decrease the memory consumption of a model. However, an unguided reduction of parameters can lead to a significant decrease in accuracy. It is crucial to carefully decide where to modify and what type of modification should be made to optimize the model architecture. As such, we based our optimization of a deep neural network architecture on well-studied design considerations described in \cite{iandola2016squeezenet} that aim to reduce the model size while maintaining accuracy. These optimizations methods were implemented on Xception \cite{chollet2017xception} as base architecture to study the feasibility of the approach.    

We evaluated our proposed model with two experiments: image classification on Caltech-101 and PCB defect detection using binary classification. The results showed an improvement in the memory and utilization inference time of our optimized architecture relative to the original Xception network. This optimization also did not sacrifice the accuracy of our model, which maintained a similar accuracy to the original network. An analysis our findings is presented in the following sections.

\subsection{Pareto Optimality of Optimized Architecture}
\begin{figure} 
    \centering
    \subfloat[Caltech-101 Image Classification\label{fig_pareto_caltech}]{%
    \includegraphics[width=0.5\linewidth, height=7cm, keepaspectratio]{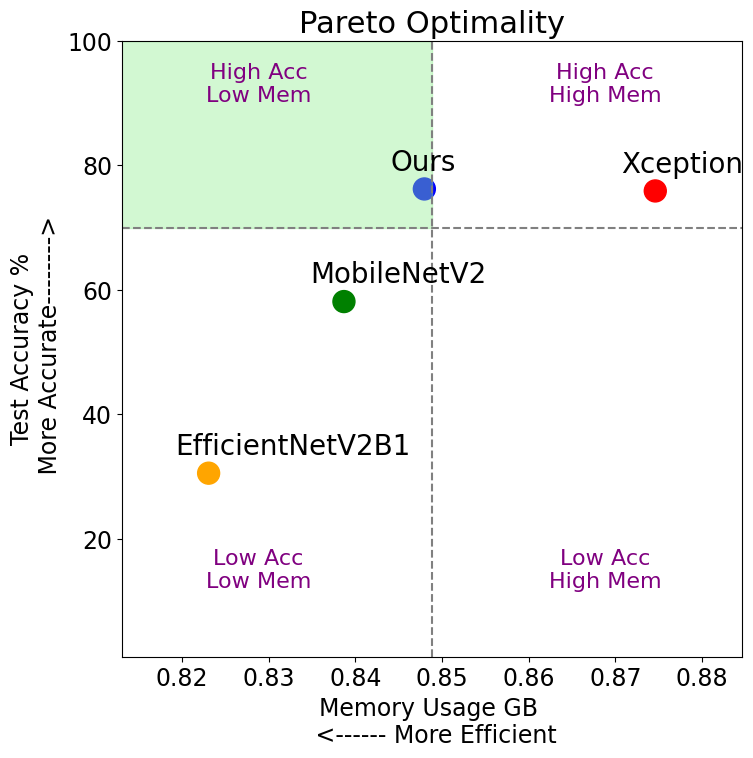}}
  \subfloat[PCB Defect Detection\label{fig_paretoPCB}]{%
        \includegraphics[width=0.5\linewidth, height=7cm, keepaspectratio]{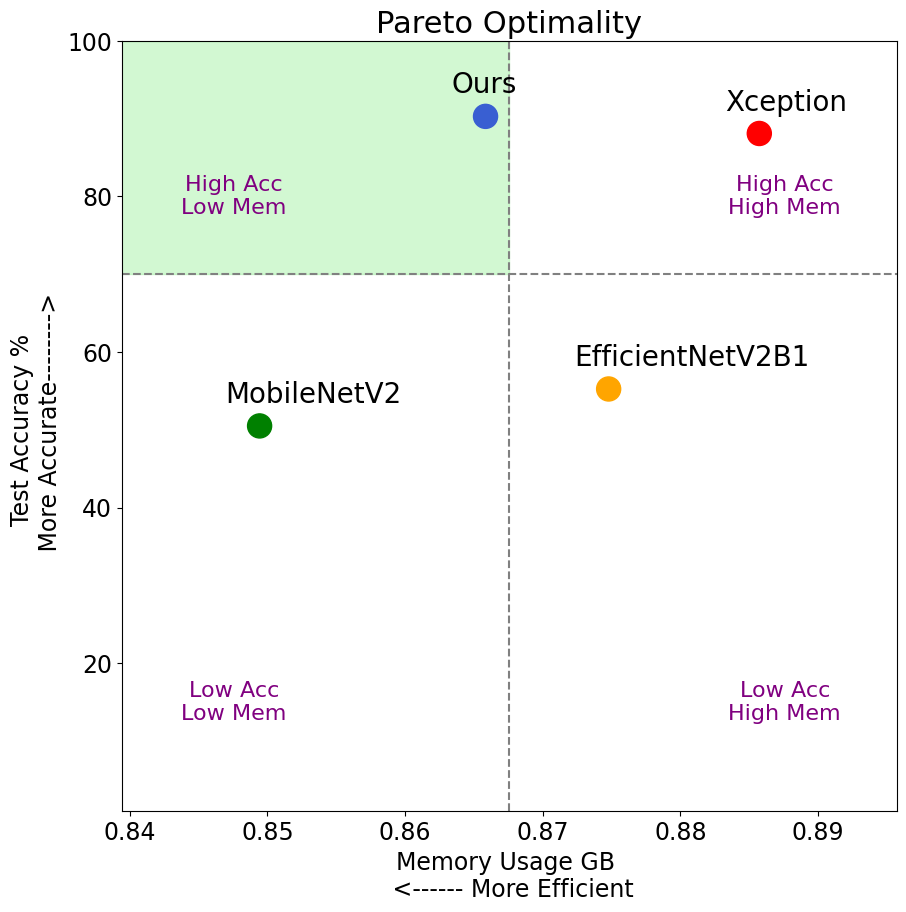}}    
  \caption{Dual-objective Analysis of Models for Accuracy and Memory Usage. In both comparisons, we define our accuracy frontier at 70\% and our memory usage frontier at the midpoint of the min and max values.}
        \label{fig:pareto_analysis}
\end{figure}

Since our work focuses on training performance, we analyze the models for accuracy and memory (as inference latency is more suitable for assessing inference performance). On a dual-objective front, our model achieves comparable accuracy to a heavyweight model (Xception) and better memory usage in both Experiment 1 and Experiment 2. An overview of the comparison is presented in Figure \ref{fig:pareto_analysis}.

In both plots, we observe that our optimized model lies within the Pareto-optimal area of the graph i.e. \textit{High Accuracy, Low Memory}, whereas the original Xception model lies on the \textit{High Accuracy, High Memory} region. On the Caltech-101 classification task (as shown in Figure \ref{fig_pareto_caltech}), the lightweight models (EfficientNetV2B1 and MobileNetV2) lie on \textit{Low Accuracy, Low Memory} region suggesting that they trade off accuracy for better hardware efficiency. The plot for the PCB defect detection task (Figure \ref{fig_paretoPCB}) shows a different observation with EfficientNetV2B1 which now has high memory usage but maintains low accuracy. This could suggest that its memory utilization scales with the size of the dataset as Experiment 2 has 22,000 images while Experiment 1 has 9,146 images. MobileNetV2 remains in the \textit{Low Accuracy, Low Memory region},  our optimized model remains in the \textit{High Accuracy, Low Memory} region, and Xception remains in the \textit{High Accuracy, High Memory} region.   

With these observations, we can conclude that our model is Pareto-optimal as it achieves both high accuracy and low memory utilization as its objectives.


\subsubsection{To Pre-train or Not?}
Wang et al.\cite{wangX2020convergence} suggested that transfer learning can facilitate training on resource-constrained devices. The results in Table \ref{tab:pcb_results} validate this notion by demonstrating an improved memory utilization of the models when using pre-trained weights. Pre-trained weights provide faster convergence than training with randomly initialized weights \cite{Hussain2019}, thereby reducing the memory requirements for training. In terms of accuracy, both our optimized model and Xception suffered a decrease in performance due to negative transfer learning. A valid reason for this the source task (Caltech-101 classification) on which the models were initially trained. If the source task significantly differs from the target task, the target performance may fail to improve or could decrease \cite{torrey2010transfer}.


From the observations drawn from Table \ref{tab:pcb_results}, we conclude that transfer learning has a positive effect on memory utilization as all the models had reduced memory utilization. However, careful considerations must be made to ensure that accuracy does not decrease such as choosing a relevant source task on which the model is trained to avoid negative transfer learning on the target task \cite{torrey2010transfer}.  

\subsection{Conclusion and Future Works}
We re-iterate our research question: \textit{Can deep learning models be optimized to facilitate training at the edge with limited resources while maintaining high accuracy with less resource consumption?} From the results and their analyses, we conclude that a suitable optimization approach can improve the hardware efficiency of existing deep learning models without sacrificing accuracy. The Pareto optimality observed in our model (as compared to the original Xception architecture) shows the success of this optimization technique. 

For our future work, we will explore a more automated design process using neural architectural search procedures with this approach and different architectural templates, thus enabling the discovery of more optimal architectural designs. We will also investigate vertical compact network design optimization methods. 



\section*{Acknowledgements}
This work was supported by the NBIF Talent Recruitment Fund (TRF2003-001). The Edge Devices used in the experiments were supported by CFI Project Number 39473 - Smart Campus Integration and Testing (SCIT Lab).

\appendix


\printbibliography[heading=subbibintoc]

\end{document}